# The Many Voices of Duying: Revisiting the Disputed Essays Between Lu Xun and Zhou Zuoren

**Xin Xie[1], Jiangqiong Li[2] and Haining Wang[3, *]**

1 Department of Information Management, Key Innovation Group of Digital Humanities Resources and Research, Shanghai Normal University, PRC

2 Department of Counseling and Educational Psychology, Indiana University Bloomington, USA

3 Department of Information and Library Science, Indiana University Bloomington, USA

**Abstract**

Lu Xun (鲁迅) and Zhou Zuoren (周作人) stand as two of the most influential writers in modern Chinese literature. Beyond their familial ties as brothers, they were also intimate collaborators during the nascent stages of their writing careers. This research employs quantitative methods to revisit three disputed essays pseudonymously published by the brothers in 1912. Our stylometric analysis uses an interpretable authorship attribution model to investigate the essays' authorship and examine the brothers' respective writing styles. Our findings suggest that 'Looking at the Country of China' (望华国篇) was authored by Lu Xun. Moreover, 'People of Yue, Forget Not Your Ancestors' Instructions' (尔越人毋忘先民之训) seems to be either predominantly authored or extensively revised by Lu Xun given its notable stylistic similarities to 'Looking at the Land of Yue' (望越篇), a piece Zhou Zuoren recognized as his own, but edited by Lu Xun. The third essay, 'Where Has the Character of the Republic Gone?' (民国之征何在), exhibits a 'diluted', mixed writing style, suggesting thorough collaboration. We offer visual representations of essay features to facilitate a nuanced and intuitive understanding. We have uncovered evidence suggesting Lu Xun's covert engagement with social issues during his purported 'silent era' and provided insights into the brothers' formative intellectual trajectories.

## 1. Introduction

*1.1 Lu Xun's Literary Achievements and Legacy*

Lu Xun (1881–1936) is one the most famous and prominent writers in modern China. While his real name is Zhou Shuren, he is best known by his pen name, which he adopted upon publishing his famous novel, *Diary of a Madman* (狂人日记), in 1918. His works cover a wide range of genres, including novels, essays, and poems, and are celebrated for their incisive social commentary, as well as their use of irony and satire.

      Lu Xun dedicated a lifetime to producing a vast collection of nonfiction works. Many essays were written with the keen observation and narrative style of a historian, criticizing societal issues of the time and relating them to the past. These works serve as a 'poetic history' of the times and society in which he lived. Although the cultural and ideological journey that Lu Xun undertook mirrors the intellectual evolution experienced by a significant number of Chinese intellectuals in the early 20th century (Zhu and Yang, 2015), Lu Xun's prose stands out for incorporating the ethos of traditional Chinese prose and elements from foreign literature. Characterized by its crisp language, flexible structure, and polished style, his writing raised the bar for the artistry of modern Chinese prose both in form and content.

---

[*] Corresponding author:
Haining Wang
Address: Luddy Hall 2115F, 700 N. Woodlawn Avenue, Bloomington, IN, USA 47408
Email: hw56@indiana.edu

As early as 1927, American scholar Robert Bartlett introduced Lu Xun in an article published in the journal *Current History*, among other modern Chinese intellectual leaders. He devoted a considerable portion of the article to praising Lu Xun's stories such as *The True Story of Ah Q* (阿Q正传), *Tempest in a Teacup* (风波), and *Diary of a Madman*, referring to Lu Xun as the 'pillar' of the New Culture Movement[1] (Bartlett, 1927). This marked the beginning of Lu Xun studies in the English-speaking world. Since then, many studies have delved into the artistic and sociological dimensions of Lu Xun's writings, often framed within the evolution of the nation's process of modernization from an imperial, colonial society. To date, Lu Xun's writings maintain significant influence as they not only attract scholars of Chinese literature, but also can be found in secondary and tertiary education textbooks across East Asia. In summary, Lu Xun commands high stature, profound impact, and lasting veneration in the literary world (Davies, 2013), and holds a place of esteem among literary critics.

*1.2 Challenges in Assembling Lu Xun's Literary Legacy*

A comprehensive collection is foundational for the in-depth reading and study of any author. However, constructing a complete collection for Lu Xun is particularly challenging due to his prolific use of pseudonyms, widely dispersed works and communications, and the existence of forgeries.

Over his literary career, Lu Xun adopted more than 150 pen names (Xu and Qin, 1988), of which 'Lu Xun' is but one, as noted earlier in this article. No one, including Lu Xun himself, has made an accurate record of his published works. These frequent changes in pen names, a tradition rooted in the practices of ancient Chinese scholars, allowed him to convey his emotions and ideals from subtly different societal perspectives. For example, he used the pen name 'Jiagan' (家干) to commemorate his experiences during his youth; 'Jiagan' implies attentiveness to family matters and an aptitude for various tasks. In his later years, the adoption of various pseudonyms primarily served as a means to circumvent the Republic of China government's censorship.

The use of pseudonyms is further complicated by the wide dispersion of publication venues and communications. Significant efforts have been made to gather these scattered works. Notable among these collections are the *Supplementary Collection to the Complete Works of Lu Xun* (Tang, 1946), its subsequent 1952 edition, the *Supplement to Lu Xun's Letters* (Wu, 1952), *Scattered Works of Lu Xun* (Editorial Team of Lu Xun Dictionary, 1979), the *Complete Collection of Scattered Works of Lu Xun* (Liu, 2001), and a 2006 supplement by Liu Yunfeng.

Several Lu Xun forgeries have surfaced over the years. For example, nine passages taken from Zhou Zuoren's 'Literary Chatters' (艺文杂话) were deceitfully attributed to Lu Xun and circulated as such (Chen, 2008). In this context, the compilation of the *Complete Collection of Lu Xun* adopted a conservative approach. As advised by Xu Guangping, Lu Xun's spouse, the editorial team worked diligently to sift the genuine from the spurious and to avoid the inclusion of questionable pieces (Hu, 1950). When certain essays closely align with Lu Xun's viewpoints, appear in the same publications, or utilize pseudonyms known to be his, deciding on their inclusion proves challenging due to the lack of definitive evidence.

**2. The Works and Their Disputes**

*2.1 The Mystery of the 'Duying' Essays*

The authorship of several essays has become a matter of debate as more pseudonymously composed works have been attributed, sometimes dubiously, to Lu Xun. In 1981, Peng and Ma happened upon four essays in *Yueduo Daily*, all penned under the pseudonym 'Duying' or simply 'Du'. Both pen names share the common element 'Du'(独), alluding to traditional Chinese classics by Zhuangzi (庄子), a Daoist philosopher, which champions the ideals of individuality and a free spirit (Li, 2016). The disputed works—'Looking at the Land of Yue' (望越篇), 'Looking at the Country of China' (望华国篇), 'People of Yue, Forget Not Your Ancestors' Instructions' (尔越人毋忘先民之训), and 'Where Has the Character of the Republic Gone?' (民国之征何在)—were all published in early 1912, during the 1911 Revolution.

These essays critically evaluated the authorities both in the Shaoxing region (in northeastern Zhejiang province, China) and at the national level. Each echoed a similar point of view: a blend of deep sorrow and indignation. While the 1911 Revolution overthrew China's last imperial dynasty at the cost of numerous lives and economic losses, the newly established government of the Republic of China fared no better in governing the country. Through a qualitative analysis of the ideas and language, Peng and Ma (1981) theorized that these pieces were Lu Xun's handiwork, suggesting that both 'Duying' and 'Du' were merely a cloak for the famed writer.

However, these essays were not included in the *Complete Collection of Lu Xun* (鲁迅全集) published in the same year (Lu, 1981). Moreover, Zhou Zuoren (1885–1967), Lu Xun's brother and himself an esteemed essayist, claimed 'Duying' as one of his own pen names in his 1970 memoirs. He explicitly mentioned that 'Looking at the Land of Yue' was his creation, with Lu Xun participating in the editing process. As a result, Zhong Shuhe, a renowned publisher and historian, incorporated these essays into the *Complete Prose Collection of Zhou Zuoren* (周作人散文全集) (Zhou, 2009). Interestingly, these very essays are missing from the collection that Zhou Zuoren edited himself (Zhou, 2002). This suggests that Zhou Zuoren, for some reason, did not hold these works in high regard and may not even have acknowledged them as his own.

*2.2 Challenges in Disentangling the Literary Legacies of the Zhou Brothers*

Zhou Zuoren was a younger brother of Lu Xun. Both of them studied in Japan: Lu Xun went in 1902, at 21, and returned to China in August 1909, with a brief return in August 1906. Zhou Zuoren accompanied Lu Xun in August 1906 and returned to China in May 1911. At the time, Zhou Zuoren looked up to and closely followed his elder brother. Concurrently, Lu Xun provided significant support to Zhou, both financially and in establishing his literary reputation. When their finances became strained, Lu Xun had to abandon his studies and return to China to earn a living so that he could continue to support Zhou Zuoren's studies.

Distinguishing between the works of Lu Xun and Zhou Zuoren is a challenging task for several reasons. Primarily, the fraternal bond between the two was not merely familial but also professional, especially during the early stages of their writing careers. To help Zhou Zuoren establish a himself in the literary world, Lu Xun published a collection he compiled under Zhou Zuoren's name, titled *A Collection of Stories from Kuaiji County* (会稽郡故书杂集). In 1909, the siblings collaborated on the translation of a series of short stories, culminating in the publication of *Foreign Short Stories* (域外小说集) Volumes I and II in Tokyo. A decade later, upon the book's reprinting, Lu Xun penned a preface, once again using Zhou Zuoren's name.

Differentiating between the brothers' works becomes even more complex given their similar intellectual and educational backgrounds, which led to notable stylistic similarities in their writings. While in Tokyo, the brothers immersed themselves in global intellectual discourses, delving into relevant Western publications such as Arthur Henderson Smith's *Chinese Characteristics*. They also exhibited a keen interest in Chinese cultural studies, drawing insights from eminent Chinese scholars such as Liang Qichao and Zhang Taiyan. In the realm of literary creation, both brothers were fervent proponents of the literary revolution. Their writings, while diverse, resonated with a similar voice that emphasized the importance of life's realities, societal concerns, and the authentic portrayal of life in literature.

**3. Literature Review**

*3.1 Review of Research on Disputed Essays*

The scholarly community has long been fascinated by essays that the brothers might have written pseudonymously. In their early writings, Lu Xun and Zhou Zuoren occasionally shared their pen names. For instance, some works authored by Lu Xun were published under Zhou Zuoren's name, and vice versa. Others, though penned by Zhou Zuoren, were credited to Lu Xun. There were also collaborative works that were casually signed under one of the brothers' names upon publication. Scholars such as Zhang (2002) have identified as many as 16 early works signed by the Zhou brothers under each other's names, including poems, translations, and book prefaces, though none of these were written under the pseudonym 'Duying'.

Peng and Ma (1981) were the first to argue that the three essays examined here, along with 'Looking at the Land of Yue', were authored by Lu Xun and could not have been the works of Zhou Zuoren. This conclusion was drawn from the similarities between these essays and Lu Xun's previous writings. Their study was inspired by Yang's (1979) and Chen's (1980) research on the style and authorship of the Duying pieces in the *Tianyi Newspaper*. However, they were not aware that Zhou Zuoren had acknowledged that 'Looking at the Land of Yue' had received Lu Xun's edits (Zhou, 1970). Peng's assertions have been echoed by several researchers, including Liu (2021), who supported claims of Lu Xun's involvement in these essays. Liu posited that, in the absence of conclusive evidence, these compositions should be viewed as collaborative efforts between the brothers. Meng (2017) hypothesized that, while Lu Xun played a leading role in the collaboration from choosing the subject matter to its refinement, Zhou Zuoren, with his superior command of English, likely furnished the requisite translations of materials written in English.

*3.2 Authorship Attribution*

While these studies offer valuable insights, they primarily rely on impressionistic interpretations of stylistic cues or qualitative analyses of thought trajectories. However, the brothers possess an ample body of work, sufficient to support a statistical analysis of their writing style. Prior research on authorship disputes has consistently found that individuals exhibit unique writing patterns, as evidenced in both controlled field studies (Baayen *et al.*, 2002) and large corpora (Narayanan *et al.*, 2012). Intuitively, when conveying a particular meaning, an author has a wide array of choices in terms of word selection, sentence structure, and rhetorical devices. Yet, despite this plethora of options, authors tend to favor specific expressions over others. Such consistent and unique use of language can be easily used for authorship attribution, which infers the likely author of a disputed text based on the quantitative analysis of lexical and syntactic features found in the text (Juola, 2006).

Consider the well-known application of authorship attribution to the *Federalist Papers*, 85 essays penned under the pseudonym 'Publius' advocating for the ratification of the United States Constitution. The attribution of twelve of these essays was long in dispute between candidate authors Alexander Hamilton and James Madison. Mosteller and Wallace (1963) identified certain function words used distinctly by the two: for instance, in their respective prior writings, Hamilton employed the word 'upon' approximately three times per thousand words, whereas Madison used it at a rate closer to one per six thousand. Since the disputed essays feature the word 'upon' at a rate significantly lower than Hamilton's typical usage, Madison emerges as the more probable author. In all, Mosteller and Wallace examined the frequencies of 30 informative function words, including 'upon', successfully attributing the 12 contested essays in the *Federalist Papers*. We observed striking similarities between the case of the *Federalist Papers* and our own, most notably the presence of two potential authors, the availability of sufficient pre-existing corpora, and the use of languages with less inflection, namely English and Chinese.

Indeed, stylometric analysis has played a pivotal role in several other debates over historical authorship, such as the contested plays between Shakespeare and Fletcher (Matthews and Merriam, 1993), the rabbinic responsa *Torah Lishmah* (Koppel *et al.*, 2007), and the Latin visions *Visio ad Guibertum Missa* and *Visio de Sancto Martino* (Kestemont *et al.*, 2015), to name a few. The similarity of those studies lies in their reliance on the distribution of function words to discern writing styles and consequently authorship. Indeed, the use of function words, or in terms of Chinese character n-grams, '虚字' or '虚词' (as Chinese does not clearly delineate idiographic units) stands as a recognized stylistic marker in authorship attribution. This method has been effectively applied across various languages, including classical Chinese. For example, Chen (1987) delved into the authorship of the last 40 chapters of *Dream of the Red Chamber* (红楼梦) by examining the distribution of function character n-grams. Higashi (2008) carried out a stylistic analysis of Ouyang Xiu's *Historical Records of the Five Dynasties*, centering on the use of function character n-grams. Likewise, Wang *et al.* (2021) underscored the significance of function character n-grams in determining authorship in Ming–Qing fiction crafted in classical Chinese.

*3.3 Function Words as Stylistic Markers*

The success of using function words as stylistic markers can be attributed to several factors. First, this class of words is prevalent in written works, accounting for a substantial proportion in English documents, e.g. 37 percent of all words in this manuscript.[2] Second, the use of function character n-grams is flexible and often subconscious, which allows them to faithfully reflect the nuances of an author's style. As Liu (2004) commented in the preface to *Analysis of Function Words* (助字辨略), the art of employing function characters captures the 'temperament' of an essay. For instance, the function characters '也' and '矣', typically used at the end of a sentence to express a definitive tone, can vary significantly between individuals as their omission does not directly alter the text's meaning. Furthermore, when a function character n-gram provides multiple alternatives, it often reflects personal preferences, as indicated by Koppel's study on synonyms (Koppel *et al.*, 2006), which encompasses function words. This is intuitive because, in extreme cases where there are no alternative synonyms available, a writer is compelled to use a character they would not normally choose. However, when multiple options are present, the writer's habitual preference is likely to emerge naturally.

Third, this category of words is generally unrelated to the topic (e.g. 'the' and 'on' in English; '和' and '况且' in Chinese). It is noteworthy, however, that several Chinese characters can function as either content-related or function words, depending on the context. For example, in the sentence '殊不知这是一尊文殊像', which translates to 'Little did they know, this is a statue of Mañjuśrī', the first '殊' is a function character meaning 'very'. However, when the same character appears in the bigram '文殊' (the name of a Buddha, Mañjuśrī), it forms a proper noun with the accompanying '文', which never performs functional roles. Therefore, when tallying function character n-grams at face value, there is a risk of inflating results since some character n-grams may be content-related.[3] Consequently, manual examination becomes essential to filter out topic-specific character n-grams. As noted by Hoover (2001), it is reasonable to include some extremely frequent words that are not function words 'under the assumption that their usage may also be unconscious.' Utilizing function and generic character n-grams can help avoid overfitting to a specific theme.

**4. Methodology**

*4.1 Research Question*

We aim to ascertain the authorship of the three disputed essays: 'Looking at the Country of China', 'People of Yue, Forget Not Your Ancestors' Instructions', and 'Where Has the Character of the Republic Gone?' Each of these essays could have been written by Lu Xun, Zhou Zuoren, or as a collaborative effort.

We frame our research question as one of authorship attribution. Our approach diverges from previous efforts that depended on documentary analysis and subjective impressions. Instead, we contrast the stylistic markers evident in the contested works with reference corpora from potential authors. The authorship attribution problem is framed as a multi-label text classification task. The model automatically determines an optimal set of parameters from the training samples (a process known as training or fitting), then uses these parameters to make predictions on samples in the test set (i.e. prediction). In classic authorship attribution setups, the training samples are single-authored. Thus, the model is most adept at inferring pieces written by a single author when trained on single-authored pieces.[4] In our scenario, however, it is possible that one or more of the essays were collaboratively composed. For instance, Zhou Zuoren acknowledged that 'Looking at the Land of Yue' underwent revisions by Lu Xun. Additionally, although no third author or significant publisher edits have been reported, such interventions cannot be completely ruled out.

First, we gently relaxed the single-author premise of classical authorship attribution by building an interpretable authorship attribution model that facilitates an intuitive understanding of the predictions. If the essays were indeed co-authored, the stylistic markers of each disputed work would fail to align well with those of either brother. Consequently, an effective classifier, employing writings solely authored by each of the brothers, would attribute a lower confidence score to collaborative works than to pieces penned by a single author. Moreover, an interpretable classifier could facilitate a nuanced *post hoc* examination of stylistic markers. For instance, in a deeply collaborative scenario, it is conceivable that features strongly associated with each author pervade the entire text. If an author's features reside within a small area and are surrounded by features indicative of the other author, it is likely that the particular author made some edits to those specific lines, especially when multiple semantically equivalent expressions exist. We devoted substantial effort to constructing an interpretable classifier, and author-specific features are visualized in the disputed essays to facilitate a straightforward interpretation of potential collaborative patterns. We continue the discussion of building an explainable stylistic classifier in Sections 4.2 and 4.3.

Second, we introduce an additional statistic to determine whether substantive editorial interventions were made to the test samples. We thereby ensure that the distribution of the test samples closely mirrors that of the training samples, thus aligning the classifier's predictions with samples of similar distribution. This statistic is computed as the total count of the most discriminative features of the brothers over the total characters of a document. We name it 'stylome density', as it is the quantity per unit ('density') of indicative features ('stylomes').

Intuitively, good, informative stylistic features recur in an individual's writing.[5] This principle might also be relevant in a collaborative context. Imagine a scenario where sentences from the corpora of Lu Xun and Zhou Zuoren are randomly combined to create a synthetic sample. The expected stylome density of this synthetic sample would correspond to a weighted sum of the densities observed in both authors' writings. However, real-world collaborations might be more complex, with both authors making compromises, leading to a 'diluted' style with a lower-than-expected stylome density. Arguably, the most distinctive stylistic features can be derived from a classifier designed to differentiate the authors' writing from a generic corpus of early 20th-century classical Chinese. Yet we opt for the authorship attribution task that differentiates the brothers' writings from each other, believing it to be a more challenging task that can extract more informative features than the former. We will demonstrate that the frequency of the most indicative features, represented by stylome density, can serve as a useful measure to determine whether a sample originates from one or both authors.

Lastly, 'Looking at the Land of Yue' serves as a reference point, given its known collaborative nature. Should the classifier's confidence and feature visualization closely align with patterns observed in 'Looking at the Land of Yue', it would suggest potential joint authorship between the brothers, especially considering the essays' proximate publication dates. (See Table 1 for publication dates.) An effective stylistic classifier distinguishing prose from the brothers should yield a relatively low score on this sample compared to other validation samples known to be written by a single author. Moreover, the stylome density of the validation and test samples should fall within a reasonable range of the stylome density distribution found in the training set, as can be assessed with straightforward hypothesis testing. We provide further details in Section 4.4.

*4.2 Corpus*

We began by assembling a corpus consisting of the unequivocal works from each author and the disputed essays. First, we gathered writings in classical Chinese that best represented the authentic styles of the authors. These works were published during the authors' lifetimes, eliminating any doubts about authorship. The chosen texts are nonfiction prose primarily published between 1903 and 1913, hence roughly contemporaneous with the disputed works published in 1912. By aligning these elements, we aimed to minimize concerns regarding potential influences on writing style due to time (Glover and Hirst, 1996), genre (Kestemont *et al.*, 2012), and register (Wang *et al.*, 2021). We further divided this collection into training and validation sets. Excluding six essays reserved for validation, the rest of the texts were used for training. We selected validation texts that covered slightly different topics from the training samples, intending to gauge the classifier's potential accuracy when confronted with unfamiliar topics. Furthermore, we used 'Looking at the Land of Yue', a known collaborative piece between the authors, as an extra validation sample. Analyzing the classifier's response to this particular sample offers insights into its behavior when confronted with collaborative works.

The three disputed essays were then gathered: one was published under the pseudonym 'Duying' and the others under 'Du'. These four essays constitute the 'test set', the authorship of which we aim to infer.

The test and validation samples, including 'Looking at the Land of Yue', and all the training samples for Zhou Zuoren were taken from Zhou (2009). Training and validation samples for Lu Xun were manually transcribed from the *Complete Collection of Lu Xun* (Lu, 2005). The sources were chosen for their acknowledged high quality. All essays are in UTF-8-coded simplified Chinese. See Table 1 for the summary of the corpus and Fig. 1 for an image of 'Looking at the Country of China' as originally published in *Yueduo Daily*.

| Author | Split | Title | Topic | Pub. Date | Length | Chunks |
|---|---|---|---|---|---|---|
| LX | Train | Lessons from the History of Science (科学史教篇) | history, science | 1908.06 | 7,032 | 7 |
| | Train | On the Aberrant Development of Culture (文化偏至论) | culture, politics | 1908.08 | 8,556 | 7 |
| | Validation | On Radium (说鈤) | science | 1903.10.10 | 3,094 | 4 |
| | Validation | On the Power of Mara Poetry (摩罗诗力说) | literary, politics | 1908.02–03 | 23,779 | 22 |
| ZZ | Train | Preface to Midst the Wild Carpathians (《匈奴奇士录》序) | literary | 1908.05 | 627 | 1 |
| | Train | Preface to Charcoal Drawing (《炭画》序) | literary | 1909.03 | 258 | 1 |
| | Train | Preface to The Lost History of Red Star (《红星佚史》序) | literary, history | 1907.03 | 1,145 | 1 |
| | Train | Preface to The Yellow Rose (《黄蔷薇》序) | literary | 1911.01 | 786 | 1 |
| | Train | A Brief Discussion on Fairy Tales (童话略论) | literary, history | 1913.11.15 | 3,093 | 4 |
| | Train | A Study on Fairy Tales (童话研究) | literary, history | 1913.08 | 5,083 | 6 |
| | Validation | Preface to Qiucao Garden Diary (《秋草园日记》序) | history | 1905.01.06 | 178 | 0.7 |
| | Validation | An Addendum to Yisi Diary (乙巳日记附记一则) | culture, history | 1905.03.30 | 63 | 0.3 |
| | Validation | A Glimpse of Jiangnan Examinees (江南考先生之一斑) | history | 1903.09.13 | 147 | 0.5 |
| | Validation | Plight and Broil in a Steamboat (汽船之窘况及苦热) | history | 1903.09.13 | 144 | 0.5 |
| ZZ & LX | Validation* | Looking at the Land of Yue (望越篇) | history, politics | 1912.01.18 | 750 | 1 |
| Duying | Test | Looking at the Country of China (望华国篇) | history, politics | 1912.01.22 | 750 | 1 |
| Du | Test | People of Yue, Forget Not Your Ancestors' Instructions (尔越人毋忘先民之训) | history, politics | 1912.02.01 | 308 | 1 |
| | Test | Where Has the Character of the Republic Gone? (民国之征何在) | history, politics | 1912.02.02 | 345 | 1 |

Table 1 Overview of the corpus

We carefully replaced direct quotations and missing characters with a sequence of a chosen mask token (⊠), repeated a corresponding number of times. This approach provides a precise denominator for calculating relative frequencies and eliminates stylistic noise extraneous to the author. Lengthier training samples were segmented into roughly 800-character chunks respecting paragraph breaks. Brief validation samples were concatenated (see 'Chunks' in Table 1). Chunking and merging in this fashion promotes statistical stability and hence aids in constructing a more robust stylistic classifier.

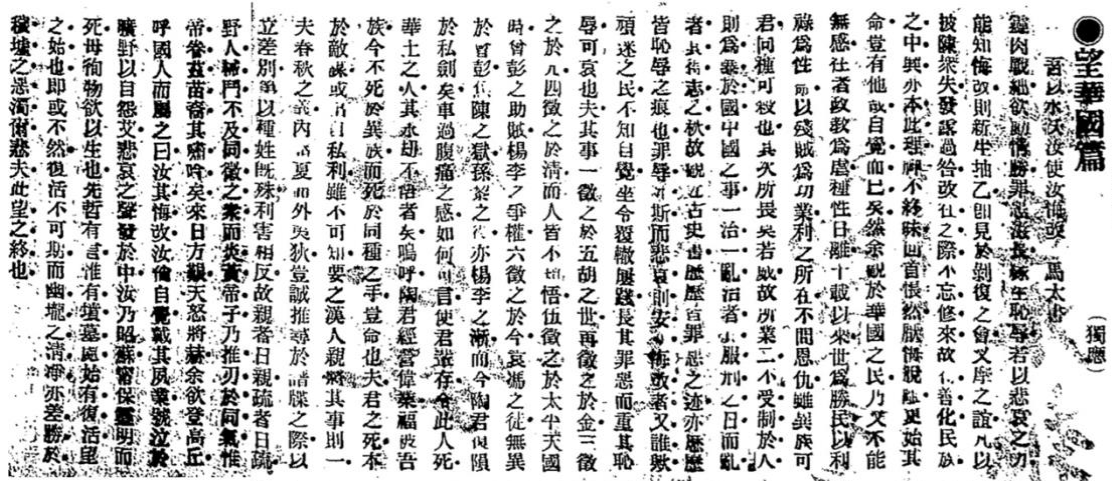

Figure 1 'Looking at the Country of China' as published in *Yueduo Daily* on January 22, 1912

*4.3 Features*

We employed the relative frequencies of function character n-grams as the stylistic markers for characterizing the writing styles of the brothers. We started with a curated list featuring more than eight hundred function character n-grams from both classical and modern Chinese, transcribed from dictionaries. After eliminating n-grams not found in the training set, our list comprised 321 n-grams. We further refined this list to identify a leaner subset that demonstrated superior performance on the validation set.

We performed two phases of recursive feature elimination, which selects distinguishing features by continually pruning the least important ones until reaching the desired number of features. The importance of a feature is determined by an underlying classifier. We selected an L2-regularized logistic regression model with regularization strength 1.0 (i.e. λ=1.0; see detailed explanation in Section 4.4) as our classifier. Instead of directly setting the minimal number of features to a sufficiently small number (i.e. 1) and performing one large search, we started with a minimum of 100 features and decreased in increments of 5. We did this because features pruned in previous iterations may be helpful in subsequent iterations, potentially due to less discriminative features heavily correlated with the pruned yet useful feature also being removed. Performing recursive feature elimination with varying minimum numbers of features can yield richer information about the most distinguishing subset. In each round, we performed leave-one-out cross-validation on the training set and recorded the performance on the validation set.

We observed that with roughly forty n-grams, the validation accuracy started to dip below an optimum of approximately ninety percent. We then conducted a more granular search within the range of 40 n-grams, specifically from 45 to 25, decreasing the minimum number of features by one each time. Eventually, we identified an optimal set of 31 highly distinguishing n-grams (27 unigrams and 4 bigrams; see Table 2) that could achieve 93.3% validation accuracy. We manually inspected the list to see whether any function character n-grams carried significant semantics. As it turns out, '诚' (honest), '光' (light), '本' (root or notebook), and '自然' (nature) are well known for their noun usage, and '随' is often employed as a verb, meaning 'follow'. However, these are generic enough and deemed not to be specific to any particular topics in our corpus. Therefore, all 31 features are considered when ascribing the authorship of the disputed essays.

*4.4 Model*

We utilized the same standard logistic regression in the main experiment as we did during the feature selection process. This model was selected for its simplicity and interpretability. The prediction from the model is

$$\hat{y} = \sigma(XW + w_0), where\ \sigma(z) = \frac{1}{1 + e^{-z}} \qquad (1)$$

where $X$ is the matrix of training data with size $m \times n$, $m=31$ is the number of training samples and $n=31$ the relative frequency of the chosen features; $W$ the vector of weights with size $n \times 1$, $w_0$ the bias term; $\hat{y}$ of size $m \times 1$. Each element of $\hat{y}$, $\hat{y}^{(i)}$ is the predicted probability of the corresponding sample belonging to the positive class, i.e. written by Lu Xun. Please note that we vectorize features based on their relative frequency, which is computed by dividing the number of occurrences by the text length in characters. Each feature is standardized to have zero mean and unit variance using statistics derived from the training set before being fed into the algorithm. The normalization of rows and standardization of columns aid in the development of robust models and enable a straightforward interpretation of feature weights.

The objective of the model is to determine the best weights (i.e. $W$ and $w_0$) using the training data. To improve the generalizability of the model, we incorporate L2 normalization to the weights as a penalty against large scales. Specifically, our logistic regression is optimized using the following loss function:

$$\frac{1}{m}\sum_{i=1}^{m}\left[y^{(i)}\log(\hat{y}^{(i)}) + (1-y^{(i)})\log(1-\hat{y}^{(i)})\right] + \frac{\lambda}{2m}\sum_{j=1}^{n}w_j^2 \qquad (2)$$

*4.5 Hypothesis Testing on Stylome Density*

We chose the most distinguishing 31 features from the logistic regression (see Table 2) and calculated the stylome density of each sample in all three splits. For each of the three testing samples, we pose the following hypotheses:

$H_0$: *The stylome density of the test sample is drawn from the same distribution as the training and validating data.*[6]

$H_a$: *The stylome density of the test sample is not drawn from the same distribution as the training and validating data.*

The empirical distribution of stylome density, as observed in the training set, closely follows a normal distribution, with mean 0.12 and standard deviation 0.02.[7] For the purposes of our analysis, we set the significance level at the conventionally accepted threshold of 0.05.

**5. Findings and Conclusions**

*5.1 Model Examination*

*Performance* Our standard logistic regression achieves a 93.3% accuracy on the single-authored validation samples. This high validation accuracy indicates the model's generalizability when applied to the three disputed writings. We will address its prediction on the collaborative validation sample, 'Looking at the Land of Yue', in Section 5.4.

*Collinearity* We checked the collinearity of the 31 features by examining their correlation matrix (Fig. 2). It appears that the pairwise correlations between features are weak to moderate, with correlation coefficients generally smaller than 0.6, except for the diagonal values. Such weak and moderate correlations may suggest that the authors exhibit unique usage patterns of the function n-grams, yet these patterns are not strongly interrelated. This could offer a finer-grained feature space to distinguish different authors.

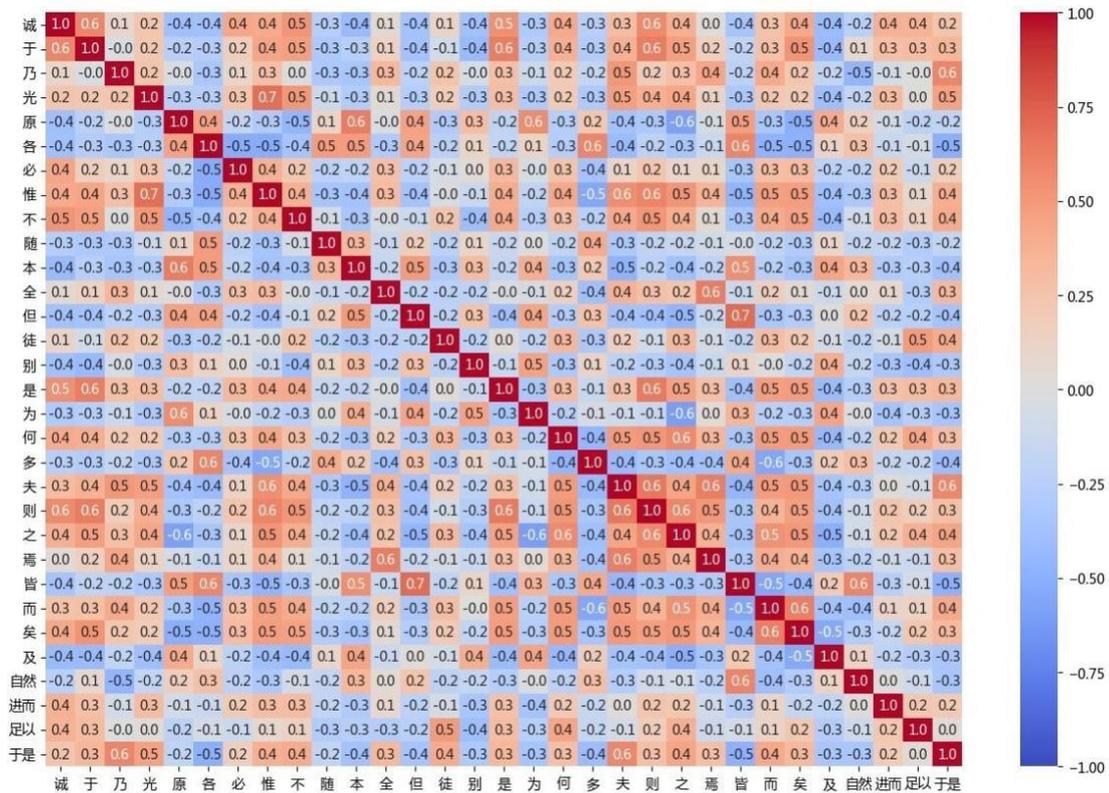

Figure 2 Correlation matrix of the feature space used in the logistic regression

*Interpreting the Weights* The weight associated with each feature is listed in Table 2. We assigned a code of 1 to samples written by Lu Xun and 0 to those by Zhou Zuoren. Consequently, features with positive weights (listed in descending order in the left column) support Lu Xun's authorship, while features with negative weights indicate Zhou Zuoren as the author.

We can intuitively understand these weights by comparing them with the appearance of the features per thousand characters in both authors' training samples. Consider the feature '则', which appears approximately nine times per thousand characters in texts written by Lu Xun, about double its frequency in Zhou Zuoren's prose (ca. five times per thousand characters). As a result, the model assigns '则' a positive weight, thus increasing the likelihood of a document being attributed to Lu Xun when '则' is present. Similarly, '各' is a character that suggests Zhou Zuoren as the author. The frequency of '各' in Zhou Zuoren's writing is twenty times greater than in Lu Xun's, even though it is a less frequently used character overall. Other function character n-grams may not be included because they are less discriminating. For instance, both authors use '无论' about 0.1 times per thousand characters. Another possible reason is that some n-grams do not contribute to a better decision, potentially due to collinearity with the selected features.

The bias term (i.e. $w_0$) has a value of 0.59. This indicates that the model would assign the authorship of the document to Lu Xun with a probability of approximately 64% when discriminative features used in the model are not present, according to Equation 1. In other words, the model leans towards attributing the authorship to Lu Xun over Zhou Zuoren when there are insufficient authorial cues present. Therefore, we must interpret the probabilities cautiously.

| Feature | Weight | Freq. per 1,000 (LX) | Freq. per 1,000 (ZZ) | Feature | Weight | Freq. per 1,000 (LX) | Freq. per 1,000 (ZZ) |
|---|---|---|---|---|---|---|---|
| Constant | 0.59 | - | - | 本 | −0.33 | 1.92 | 4.62 |
| 之 | 0.4 | 40.31 | 26.98 | 及 | −0.31 | 0.77 | 3.80 |
| 诚 | 0.35 | 0.96 | 0.00 | 别 | −0.28 | 0.26 | 0.91 |
| 而 | 0.31 | 15.32 | 8.96 | 原 | −0.27 | 0.58 | 2.81 |
| 惟 | 0.29 | 2.63 | 0.36 | 各 | −0.26 | 0.06 | 1.18 |
| 矣 | 0.28 | 3.91 | 0.81 | 为 | −0.24 | 9.68 | 15.03 |
| 是 | 0.26 | 4.93 | 1.72 | 多 | −0.24 | 2.56 | 5.16 |
| 于是 | 0.26 | 1.35 | 0.18 | 但 | −0.22 | 0.06 | 1.45 |
| 足以 | 0.25 | 0.64 | 0.18 | 皆 | −0.21 | 0.77 | 3.17 |
| 必 | 0.25 | 2.05 | 0.54 | 自然 | −0.20 | 0.38 | 1.27 |
| 何 | 0.24 | 1.92 | 0.18 | 随 | −0.19 | 0.06 | 0.36 |
| 徒 | 0.24 | 0.70 | 0.18 | | | | |
| 焉 | 0.23 | 1.22 | 0.18 | | | | |
| 不 | 0.23 | 13.27 | 6.97 | | | | |
| 乃 | 0.22 | 3.08 | 1.90 | | | | |
| 于 | 0.22 | 12.88 | 8.15 | | | | |
| 则 | 0.21 | 8.59 | 4.71 | | | | |
| 进而 | 0.21 | 0.32 | 0.00 | | | | |
| 全 | 0.2 | 0.83 | 0.18 | | | | |
| 光 | 0.19 | 0.58 | 0.00 | | | | |
| 夫 | 0.18 | 2.44 | 0.45 | | | | |

Table 2 Feature weights and each author's usage frequency

*5.2 Examining Stylome Density of Test Samples*

The collaborative work 'Looking at the Land of Yue' has a stylome density of 0.13, which closely aligns with the average stylome density observed in known single-authored pieces. This finding confirms our hypothesis that the stylome density of collaboratively written essays also originates from the distribution of single-authored pieces.

      Upon examining the test samples (see Table 3), there is insufficient evidence to reject the null hypothesis, which suggests that the stylome density of the test samples is drawn from the same distribution as that of the training and validation datasets. As a result, we concluded that the essays were either solely or jointly authored by the two brothers, without substantive contributions from an external third party.

      We observed a marginally insignificant p-value of 0.07 for an edge case with 'Where Has the Character of the Republic Gone?'. This indicates that the stylome density of this specific sample slightly deviates from those in the training and validation sets. Such a discrepancy might result from moderate editorial changes, a fusion of individual styles during collaboration, or other unforeseen factors.

| Essay | Split | Z score | P value (two-sided) |
|---|---|---|---|
| Looking at the Land of Yue | Validation (Collaboration) | 0.62 | 0.53 |
| People of Yue, Forget Not Your Ancestors' Instructions | Test | 0.4 | 0.69 |
| Looking at the Country of China | Test | 1.59 | 0.11 |
| Where Has the Character of the Republic Gone? | Test | −1.8 | 0.07 |

Table 3 Summary of stylome density for the disputed essays

*5.3 'Looking at the Country of China' Was Written by Lu Xun*

Our model predicts with a high probability (0.984) that the author of 'Looking at the Country of China' is Lu Xun. The evidence is also clearly demonstrated in Figure 3, where features indicative of Lu Xun are highlighted in red, while those suggestive of Zhou Zuoren are rendered in gray. The characters' shading corresponds to their significance. Most of the marked features are in red, notably including '之', the strongest stylistic marker of Lu Xun, which appears in his writings approximately 40 times per thousand characters.

A function character n-gram that has more synonyms is more likely to reveal an author's personal preference (Koppel, 2006). We noted 40 instances of '之' in the essay; such a high frequency of use makes it unlikely that it was written by Zhou Zuoren. Some instances could not be edited without altering the meaning, (e.g. '君之死' and '利之所在'). However, a significant number of '之' appearances seem to be stylistically chosen. The '之' in '民族之中兴' is a clear example, in which the character appears to have been intentionally chosen to disrupt a series of four-character parallel sentences. This technique gives the sentence a sharp rhythm, guiding the reader's attention to ponder the question, 'How can the nation be revitalized?' instead of merely following the author's narrative. Moreover, the six instances of '之' in '其事一征之于五胡之世，再征之于金，三征之于元，四征之于清，而人皆不知悟；五征之于太平天国时' could be removed without altering the semantics. However, the author chose to use them, further suggesting Lu Xun's authorship.

We also observed minor contrary evidence. The character '本' (a preposition meaning 'according to') is one of the characteristic stylistic markers of Zhou Zuoren. The '本' in '亦本此理' has many alternatives, such as '循', '藉', and '据', which could easily replace '本'. However, the author chose the character which is more frequently seen in Zhou Zuoren's writings. Therefore, we cannot completely rule out the possibility that Zhou Zuoren, or another editor, might have made minor changes to the essay. It might be the case that this style emerged when Lu Xun composed the piece under an atypical mood or setting; perhaps it was the result of his brief indignation when he witnessed the true nature of the revolutionaries and saw his hopes dashed.

On balance, based on model predictions and manual inspection of stylistic features, we believe that 'Looking at the Country of China' was written by Lu Xun.

灵肉战绁，欲动情胜，罪恶滋长，缘生耻辱。若以悲哀之力，能知悔改，则新生抽乙，即见于剥复之会，又摩之以谊，凡以披陈众失，发露过咎，改往之际，不忘修来，故待善化。民族之中兴，亦本此理。神不终昧，回首怅然。厌悔脱离，更始其命，岂有他哉，自觉而已矣。

然余观于华国之民，乃又不能无感。往者政教为虐，种性日离，千载以来，世为胜民，以利禄为性命，以残贼为功业。利之所在，不问恩仇，虽异族可君，同种可杀也。其次所畏莫若威。故所业二，不受制于人，则为暴于国。中国之事，一治一乱。治者其服刑之日，而乱者其得志之秋。故观近古史书，历历皆罪恶之迹，亦历历皆耻辱之痕也。罪辱如斯，而悲哀则安在，悔改者又谁欤？顽迷之民，不知自觉，坐令覆辙屡践，长其罪恶而重其耻辱，可哀也夫！其事一征之于五胡之世，再征之于金，三征之于元，四征之于清，而人皆不知悟；五征之于太平天国时曾彭之助贼，杨李之争权；六征之于今，袁冯之徒无异于曾彭，焦陈之狱，孙黎之𠀾，亦杨李之渐。而今陶君复殒于私剑矣，车过腹痛之感，如何可言！使君辈存，令此人死。华土之人，其永劫不悟者矣。

呜呼！陶君经营伟业，福被吾族。今不死于异族，而死于同种之手，岂命也夫！君之死，本于敌谋，或出自私利，虽不可知，要之汉人亲将其事则一。夫《春秋》之义，内诸夏而外夷狄，岂诚推寻于谱牒之际，以立差别？第以种姓既殊，利害相反，故亲者日亲，疏者日疏。野人械斗不及同征之众，而炎黄帝子，乃推刃于同气。惟帝眷兹苗裔，其啸吟矣。

来日方艰，天怒将赫。余欲登高丘，呼国人而属之曰：汝其悔改！汝倘自觉，戴其凤业，号泣于旷野以自怨艾，悲哀之声发于中，汝乃昭苏。宁保灵明而死，毋徇物欲以生也。先哲有言：惟有坟墓处，始有复活，望之始也。即或不然，复活不可期，而幽垄之清净，亦差胜于秽墟之恶浊尔。悲夫！此望之终也。

Figure 3 Visualization of features in 'Looking at the Country of China' supporting each author

*5.4 'People of Yue, Forget Not Your Ancestors' Instructions' Mirrors the Collaborative Approach the Brothers Took in 'Looking at the Land of Yue'*

'People of Yue, Forget Not Your Ancestors' Instructions' is predicted to be written by Lu Xun with a probability of 0.833, notably lower than model confidence for 'Looking at the Country of China'. This probability corresponds to five-to-one odds of being written by Lu Xun versus Zhou Zuoren, should the essay be solely written. Considering the possibility of collaboration, we notice a surprisingly similar model confidence (0.856) for 'Looking at the Land of Yue', a work Zhou Zuoren admitted had received Lu Xun's edits.

In the essay 'People of Yue, Forget Not Your Ancestors' Instructions,' the character '之'—a distinctive feature of Lu Xun's prose—appears with notable frequency. Out of its 13 occurrences in the brief essay, two instances stand out for their artistic selection. For instance, the phrase '嗟尔越之人，其敢忘先民之训乎' aptly encapsulates the essence of the essay's title. In the latter segment of this phrase, the use of '之' serves as a modifier between '先民' and the pivotal term '训,' establishing a possessive relationship, rendering it essential. Conversely, in the earlier segment, the term '越之人' diverges from the title's '越人'; here, the inclusion of '之' is not strictly obligatory but is deliberately chosen. A comparable deployment of this character is evident in the clause '越之人实有力焉'.

While Lu Xun's stylistic influence is predominant, we discern elements emblematic of Zhou Zuoren's style, notably in the employment of the word '为'. The usage of '为' in '尔幸为越雪之' could be replaced by '替' or '与', elucidating the beneficiary of the action within the sentence. This resonates with the instances of '孰为决之' and '吾为此惧' found in 'Looking at the Land of Yue'. Furthermore, in both '为报仇雪耻之乡耶？为藏垢纳污之地耶？' from the current essay and '为善为恶' from 'Looking at the Land of Yue', the term '为' might be substituted with '或' to signify an alternative or choice. However, the consistent selection of this term suggests a deliberate stylistic or habitual inclination.

There exists persuasive external evidence that also warrants attention. Both 'Looking at the Land of Yue' and 'People of Yue, Forget Not Your Ancestors' Instructions' center on the land of Yue, the ancestral homeland of the brothers. The former delves into the national character of the society, tracing over two millennia of monarchy, and articulates apprehensions regarding the prospective rejuvenation of Yue and the broader nation. The latter, the essay under scrutiny, proffers a prudential observation rooted in Yue's contemporary political milieu. It becomes manifest that the later essay is an organic continuation of its predecessor. From a documentary vantage point, Lu Xun recurrently integrated a line from this essay into his subsequent compositions: '会稽乃报仇雪耻之乡,非藏垢纳污之地', attributed to the Ming Dynasty politician Wang Siren (王思任, 1574–1646). Yet Wang's original phrasing commences with '*越乃报仇雪耻之国*', not '*会稽乃报仇雪耻之乡*', despite their synonymous implications (Peng, 1981). The likelihood of different authors consistently using the same quote and making the same referencing mistakes is slender, supporting the proposition that this essay either came directly from Lu Xun's hand or, at the very least, bears his distinctive touch.

先民有言：会稽乃报仇雪耻之乡，非藏垢纳污之地。盖越自勾践以来，遗风未泯，士尚气节。中世遭种族之变，苦心积虑，不忘报复。绍兴一郡，遂屡为诸夏君民最后诀别地，虽曰不祥，亦烈矣哉！

北房乱华，十年经营，而后报之，越之人实有力焉。生荣死哀，天下昭见。呜呼，亦越之光矣。

虽然，人生于忧患而死于安乐。东南半壁，方脱房系，而内讧频闻，形同割据。近传台绍诸郡，亦谋分立，虽曰流言，虑非佳兆。嗟尔越之人，其敢忘先民之训乎！尔为福于国，民未尝不尔感；尔为毒于国，民亦不能不尔怨也。臣房之耻，尔幸为越雪之；今尔偾事，又令谁为尔雪此耻乎？

呜呼！於越古国，而今而后，为报仇雪耻之乡耶？为藏垢纳污之地耶？危乎危乎！虽然，往者不可谏，来者犹可追也。

Figure 4 Visualization of features in 'People of Yue, Forget Not Your Ancestors' Instructions', the special validation sample collaboratively written by Zhou Zuoren and Lu Xun

盖闻之，一国文明之消长，以种业为因依，其由来者远，欲探厥极，当上涉于幽冥之界。种业者，本于国人彝德，驸以习惯所安，宗信所仰，重之以岁月，积渐乃成，其期常以千年，近者亦数百岁。逮其宁一，则思感咸通，之为公意，虽有圣者，莫能更赞一辞。故造成种业，不在上智，而在中人；不在生人，而在死者。二者以其为数之多与为时之永，立其权威，后世子孙承其血胤者亦并袭其感情，发念致能，莫克自外，唯其坐绍其业而收其果，为善为恶，无所撰别，遗传之可畏乃如是也。

盖民族之例，与他生物同。大野之鸟，有翼不能飞；冥海之鱼，有目不能视；中落之民，有心思材力而不能用。习性相重，流为种业，三者同然焉。中国受制于满洲，既二百六十余年，其局促伏处专制政治之下者，且二千百三十载矣。今得解脱，会成共和，出于幽谷，迁于乔木，华夏之民，孰不欢欣？顾返瞻往迹，亦有不能不惧者。其积染者深，则更除也不易。中国政教，自昔皆以愚民为事，以刑戮慑俊士，以利禄招黠民，益以儒者邪说，助张其虐，二千年来，经此淘汰，庸愚者生，佞捷者荣，神明之种，几无孑遗。种业如斯，其何能臧？历世忧患，有由来矣。

今者千载一时，会更始之际，予不知华土之民，其能洗心涤虑以趣新生乎？抑仍将忺忺倪倪以求禄位乎？于彼于此，孰为决之？予生於越，不能引以观其变，今唯以越一隅之为征。当察越之君子何以自建，越之野人何以自安；公仆之政何所别于君侯，国士之行何所异于臣妾。凡兹同异，靡不当详，国人性格之良窳，智虑之蒙启，可于是见之。如其善也，斯於越之光，亦夏族之福；若或不然，利欲之私，终为吾毒，则是因果相循，无可诛责，唯有撮灰散顶，诅先民之罪恶而已。仲尼《龟山操》曰："☒☒☒☒，☒☒☒☒，☒☒☒☒，☒☒☒☒。"今瞻禹域，乃亦惟种业因陈为之蔽耳。虽有斧柯，其能伐自然之律而夷之乎？吾为此惧。

Figure 5 Visualization of features in the known collaborative work 'Looking at the Land of Yue' supporting each author

*5.5 'Where Has the Character of the Republic Gone?' Reflects a Thorough Collaborative Effort*

The essay entitled 'Where Has the Character of the Republic Gone?' exhibits a style distinct from either of the authors or the recognized collaborative pattern evident in 'People of Yue, Forget Not Your Ancestors' Instructions'. The model's confidence in attributing the essay to Lu Xun is notably low, standing at 0.60. Recall that the model has an inherent bias term of 0.59. This equates to a probability of 0.64 for a document being attributed to Lu Xun even in the absence of discriminative stylistic indicators. It suggests that the model lacks awareness of the particular linguistic pattern found in 'Where Has the Character of the Republic Gone?' in its training data.

  Given the marginally insignificant p-value from the stylome density test, we exercise caution in asserting significant external edits or attributing the piece to a third author. The overall stylistic tone of the essay seems somewhat atypical when compared to other writings by the brothers, whether individually penned or collaborative. Notably, a sequence of over seventy characters in the essay, spanning from '更通观全局' to '酒资少亦十角', lacks any discernible stylistic markers characteristic of either author.

  Upon manual examination of linguistic features, the essay does not strongly align with the distinctive styles of either Lu Xun or Zhou Zuoren. For instance, the character '之' is conspicuously less frequent than in the other two essays under scrutiny. There are instances where the inclusion of '之' would have been fitting. The phrase '使我夏民', for example, parallels '华夏之民' from 'Looking at the Land of Yue' and '华国之民' from 'Looking at the Country of China'. Structurally, it might have been rendered as '使我夏之民', echoing the structure of '越之人' in 'People of Yue, Forget Not Your Ancestors' Instructions'. The choice to eschew this formulation diminishes the probability of Lu Xun being the sole author. While certain stylistic markers hint at Zhou Zuoren's influence, such as the use of '皆' (meaning 'all') in '皆有平亭' and '此皆彰彰在人耳目者', alternative expressions like '均' and '尽' could have been employed. However, given the overall diluted stylistic signal, sole authorship by Zhou Zuoren is not robustly supported.

  In light of these observations, we are inclined to conclude that 'Where Has the Character of the Republic Gone?' may be the result of a deep collaboration, wherein the distinct styles of the authors melded.

在昔满人主国，淫威孔肆，法以意造，政以贿成，使我夏民生命财产无所保入，其罪上通于天。夫惟异族专制为然，然满人亦以是自殒其命。革命既兴，种族政治，皆有平亭，旧日之淤，不难尽去，而夷考其实，信乎？否乎？

昔秋女士被逮，无定谳，遽遭残贼，天下共愤，今得昭复。而章介眉以种种嫌疑，久经拘讯，亦罪无定谳，而议籍其家。自一面言之，可谓天道好还；又一面言之，亦何解于以暴易暴乎！此矛盾之一例也。更统观全局，则官威如故，民瘼未苏。翠舆朝出，荷戈警跸；高楼夜宴，倚戟卫门；两曹登堂，桎梏加足；雄师捉人，提耳流血。保费计以百金，酒资少亦十角。此皆彰彰在人耳目者，其他更何论耶！

呜呼！昔为异族，今为同气；昔为专制，今为共和。以今较昔，其异安在？由今之道，无变今之俗。浙东片土，固赫然一小朝廷也。异于昔者，殆在是耶！

Figure 6 Visualization of features in 'Where Has the Character of the Republic Gone?' supporting each author

## 6. Discussion

*6.1 Dissecting Stylistic Nuances*

Lu Xun's distinctive authorial style is characterized by his adept use of high-frequency function characters such as '之', '而', '不', '则', '是', and '矣'. To illustrate, consider the character '矣', typically positioned at a sentence's conclusion. It is used to affirm, as demonstrated in '岂有他哉，自觉而已矣'; express exclamation, as in '而今陶君复殒于私剑矣'; or convey speculation, as illustrated by '华土之人，其永劫不悟者矣'. Another example is '夫'. When prefacing a sentence, it often introduces a discourse or interpretation, as seen in '夫《春秋》之义，内诸夏而外夷狄'; when concluding, it can evoke an exclamatory tone, as in '今不死于异族，而死于同种之手，岂命也夫！'. The judicious employment of these auxiliary terms augments the essay's rhythmic cadence, allowing the author's sentiments to resonate profoundly.

  Additionally, Lu Xun exhibits a preference for certain adverbs and conjunctions, such as '诚', '而', '足以', and '进而'. For instance, the term '乃', when serving as a conjunction, can denote consequence, as in '然余观于华国之民，乃又不能无感', or highlight an unforeseen turn of events, as in '而炎黄帝子，乃推刃于同气'. The deployment of these function characters elucidates the logical interconnections between clauses, steering readers through the essay's argumentative trajectory. This crafts a nuanced yet cogent aesthetic, emphasizing the author's discerning and methodical approach. Conversely, Zhou Zuoren exhibits a more restrained use of the aforementioned auxiliary terms. This is evident in a portion of 'Looking at the Land of Yue' (i.e. from '中国政教' to '几无孑遗'). This 70-character passage is sparse in conjunctions, trading a degree of readability for conciseness.

  In a comprehensive examination of the quartet of essays, it becomes evident that 'Looking at the Country of China,' 'People of Yue, Forget Not Your Ancestors' Instructions,' and 'Where Has the Character of the Republic Gone?' share a great similarity in sentiment. These compositions recurrently utilize emotive lexemes such as '呜呼', '乎', and '耶'. The interplay of exclamatory and declarative structures underscores the author's profound sorrow and indignation. Conversely, the earliest of the four essays, 'Looking at the Land of Yue', adopts a more tempered tone, probably reflecting the author's optimism and expectation for the emergence of a new society and the transformation of the national character shortly after the 1911 Revolution, as expressed in '如其善也，斯於越之光，亦夏族之福，' which implies great hope.

  Our analyses indicate that Lu Xun either authored or played a significant role in shaping these three essays. Considering that Zhou Zuoren's memoirs solely reference 'Looking at the Land of Yue' and neglect to mention the other three works (Zhou, 1970), we postulate that Lu Xun's distinctive literary style is permeated with fervor and emotional intensity—a conjecture congruent with our preceding dissection of the essays' linguistic nuances.

*6.2 Beyond the Pseudonym*

Our findings and conclusions on the disputed essays provide a crucial addition to prior research related to works attributed to 'Duying'. Beyond the pseudonym, our research not only provides the missing piece of Lu Xun's creative evolution but also bridges his intellectual journey.

*6.2.1 Lu Xun's Commitments in the 'Silent Era'*

While Zhou Zuoren openly admitted to using the pseudonym 'Duying,' the prevailing assumption has long been that all works under this pseudonym were exclusively his creations. However, our investigation reveals a more intricate picture: certain voices previously credited to 'Duying' bear a distinct resemblance to Lu Xun's style.

From his return to China in 1909 until the publication of his famous story *Diary of a Madman* in 1918, Lu Xun was reported to have concentrated chiefly on the collection and organization of ancient classics and regional documents. Remarkably, during this period, he was deemed to offer few commentaries on the prevailing socio-political milieu, prompting scholars to term these years Lu Xun's 'silent era'. (Qian, 2003) However, the authorship of these essays suggests that beneath the facade of silence, Lu Xun remained deeply concerned about societal issues and the revolutionary process.

In addition, these essays were crafted in classical Chinese, rendering them of unique significance in the readings of Lu Xun. Lu Xun played a pivotal role in the New Culture Movement and championed the use of vernacular Chinese: he not only advocated for but also exemplified a transformative engagement with traditional Chinese linguistic norms, aiming to free the mind from the constraints imposed by classical Chinese. Consequently, the public largely viewed him as removed from the classical Chinese tradition, with many even presuming he downplayed or opposed it. Our investigation sheds light on previously uncharted territories of Lu Xun's works in classical Chinese and delves into his distinctive stylistic nuances. Indeed, the practice of writing in the vernacular, including the four essays examined in this study, played a constructive role in shaping his writing style. These findings substantiate the notion that Lu Xun's departure from classical to vernacular Chinese concealed an underlying consistency (Qian, 2018).

*6.2.2 Bridging Lu Xun's Intellectual Journey*

Composed in the aftermath of the 1911 Revolution, the essays under study demonstrate the author's keen observations and heartfelt contemplations on the nation and its people. They play a pivotal role in Lu Xun's intellectual journey, acting as a bridge between his past and subsequent works.

Prior to the publication of *Diary of a Madman* in 1918, Lu Xun's primary literary endeavors consisted of translating foreign novels. He also produced a series of classical Chinese works around 1908, including 'On the Aberrant Development of Culture' and 'On the Power of Mara Poetry' which can be found in our corpus. These early works were devoted to the exaltation of the human spirit, a concept that laid the foundation for the intellectual framework of the young Lu Xun (Sun, 2013). Themes found in the disputed essays, such as '灵明' (spirit or consciousness) and '自觉' (self-awakening), seamlessly connect with his early works drafted in 1908, serving as a link to his earlier thoughts. These essays also present contrasts between the ancient and the present, exposing the flaws in the national character under imperial governance with a sharp tone that could barely mask the author's sorrow. These elements foreshadow the foundational themes of Lu Xun's works after 1918; hence, our findings aid in delineating the overarching trajectory of Lu Xun's intellectual evolution.

Given Zhou Zuoren's self-proclaimed indifference to revolutionary affairs following his 1911 return (Zhou, 1970), and in light of Lu Xun's evident influence in these disputed essays, it is advisable to approach these texts with caution to avoid misattributions. To our knowledge, no analogous studies on this topic have yet surfaced.

*6.3 Future Research Directions*

Our study offers a preliminary insight into the authentic identity behind the pseudonym Duying and sheds light on the collaborative dynamics between Lu Xun and Zhou Zuoren by quantitatively analyzing three disputed essays. It is also worth noting that more than a dozen essays signed by Duying appeared in the newspapers *Tianyi* and *Henan*, fronts of public opinion at the time, between 1907 and 1908. Additionally, five essays bearing Lu Xun's signature in the magazine *New Youth* (新青年) have been claimed as the work of Zhou Zuoren (Wang, 2012) yet are included in the *Complete Works of Lu Xun*. Further scrutiny of those works could deepen understanding of the voices behind Duying and further distinguish each author's unique contributions.

Lu Xun occupies an irrefutable central role in his time and beyond. While there may be writers with more refined aesthetics, the deep social insights and critical thinking present in his works are unparalleled. Many of his writings reflect a sober understanding of China's society, culture, and people, rendering Lu Xun's works profoundly relevant even today.

Our corpus and scripts for reproduction are available at https://codeberg.org/haining/the_many_voices under permissive licenses.

**Notes**

1. The New Culture Movement in China was a 20th-century intellectual and cultural reform movement. It aimed to modernize China by challenging traditional Chinese culture, advocating for democracy, science, and vernacular language, and promoting new ideas and values.
2. We reused a function word list containing 512 function words reported in Koppel *et al.* (2009).
3. Incorporating part-of-speech information could alleviate this concern. Nonetheless, a parser specifically trained for this particular language period is unavailable, due to the fact that the 1910s marked a turning point when the mainstream Chinese register transitioned from classical to vernacular. We prefer to avoid introducing additional noise and instead opt for manual scrutiny.
4. The assumption that each sample in a dataset, whether from the training or the test set, is generated from the same probability distribution and is independent of all other samples is referred to as the 'independent and identically distributed' assumption. This principle is foundational to most statistical machine learning models.
5. Admittedly, machine learning tends to take shortcuts (Geirhos *et al.*, 2020) or overfit unless it is regularized not to. Here, using the word 'good', we emphasize the importance of using the features that generalize, which are usually frequent and spread over a sample.
6. Note that we set aside the collaboratively composed essay 'Looking at the Land of Yue' to test whether it originates from the same distribution as the single-authored pieces, thereby confirming our intuition discussed in Section 5.2.
7. The assertion of normality is substantiated through an omnibus test, which takes into account both skewness and kurtosis of the distribution, by combining D'Agostino and Pearson's tests (D'Agostino, 1971; D'Agostino and Pearson, 1973). The test yields a p-value of approximately 0.43, which considerably exceeds typical significance thresholds. Thus, there is insufficient evidence to reject the hypothesis that the stylome density originates from a normal distribution. This normality is further confirmed visually via a Q-Q plot.

**Acknowledgments**

We extend our deep gratitude to Allen Riddell for his invaluable guidance and comments, to Bojue Hou for translating the titles of the works under study, to Jinzhi Zhou for her insightful discussions on Chinese function characters, and to Michael Hartwell for his expert English copyediting.

**Funding**

This work was supported by the National Social Science Foundation of China [Grant Number: 22CTQ041].

**Figure and Table Legends**

Fig. 1: The essay is written by Duying in classical Chinese, reading from top to bottom and right to left. The text features primitive punctuation markers (i.e. a dot beside a character) between clauses and sentences, which assist subsequent editors in mapping them to modern Chinese punctuation marks. Image courtesy of the Archives of Zhejiang Province, China.
Fig. 2: Features exhibit weak to moderate correlations.
Figs. 3–6: Reddish characters favor Lu Xun as the author, while gray ones indicate a preference for Zhou Zuoren. The darker the shade, the greater the absolute value of the weights for each feature. Unrecognizable characters and direct quotations are substituted with the character '⌧'.

Table 1: Works by Lu Xun (LX), Zhou Zuoren (ZZ), 'Duying', and 'Du'. Four separate validation samples from Zhou are merged into two longer texts. Lengthier essays are segmented into smaller units, with paragraph breaks duly observed. The collaborative work 'Looking at the Land of Yue' is a special validation sample.
Table 3: The reference distribution for stylome density is derived from samples in the training and validation sets (only single-authored pieces).